\documentclass[11pt,twocolumn]{article}
\usepackage{paper}

\title{Limited ability of LLMs to simulate human psychological behaviours: a psychometric analysis}

\author{Nikolay B Petrov\orcidlink{0000-0002-1305-0547}\textsuperscript{1}, Gregory Serapio-García\orcidlink{0000-0002-1890-2331}\textsuperscript{1}, Jason Rentfrow\orcidlink{0000-0002-9068-2118}\textsuperscript{1}}

\affil{\normalsize{\textsuperscript{1}Department of Psychology, University of Cambridge, Cambridge, UK}}
\date{\vspace{-5ex}}

\begin{document}

\twocolumn[
    \begin{@twocolumnfalse}
        \maketitle
        \begin{abstract}
            The humanlike responses of large language models (LLMs) have prompted social scientists to investigate whether LLMs can be used to
            simulate human participants in experiments, opinion polls and surveys. Of central interest in this line of research has been mapping out
            the psychological profiles of LLMs by prompting them to respond to standardized questionnaires. The conflicting findings of this research
            are unsurprising given that mapping out underlying, or latent, traits from LLMs' text responses to questionnaires is no easy task. To
            address this, we use psychometrics, the science of psychological measurement. In this study, we prompt OpenAI's flagship models, GPT-3.5
            and GPT-4, to assume different personas and respond to a range of standardized measures of personality constructs. We used two kinds of
            persona descriptions: either generic (four or five random person descriptions) or specific (mostly demographics of actual humans from a
            large-scale human dataset). We found that the responses from GPT-4, but not GPT-3.5, using generic persona descriptions show promising,
            albeit not perfect, psychometric properties, similar to human norms, but the data from both LLMs when using specific demographic profiles,
            show poor psychometrics properties. We conclude that, currently, when LLMs are asked to simulate silicon personas, their responses are
            poor signals of potentially underlying latent traits. Thus, our work casts doubt on LLMs’ ability to simulate individual-level human
            behaviour across multiple-choice question answering tasks.
        \end{abstract}
        \bigskip
    \end{@twocolumnfalse}
]

\begin{figure*}[tb]
    \centering
    \includegraphics
    [trim=0 0 0 0, keepaspectratio, clip, width=0.95\textwidth]
    {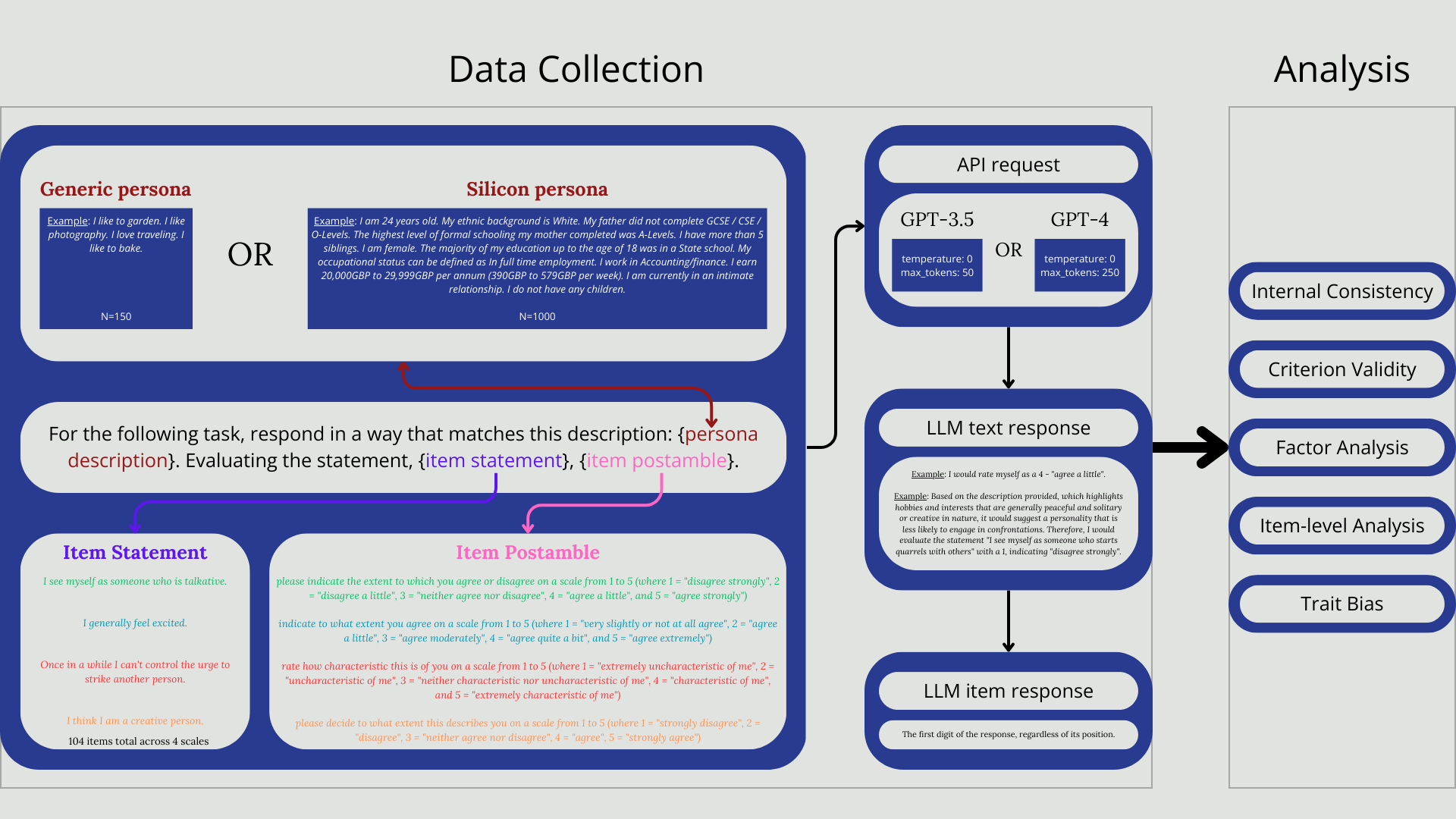}
    \caption {
        \small\textbf{Overview of the data collection and analysis processes} Each LLM (GPT-3.5 and GPT-4) is prompted using a template of a persona description and a survey item. The persona description can be either a generic one, constructed using 4-5 random sentences from the PersonaChat dataset \autocite{zhangPersonalizingDialogueAgents2018}, or a silicon one, constructed on the basis of mostly demographic information of humans from a large-scale personality survey \autocite{rentfrow_regional_2015}. Survey items that the LLM is asked to evaluate are across various personality-related constructs. The LLM text response is processed to extract a numeric response on the survey item and then further analysed.
    }
    \label{fig:method}
\end{figure*}

\section{Introduction}
\label{sec:intro}
The recent advances in AI have led to the development of large language models (LLMs). LLMs’ performance across a range of tasks matches, or even
exceeds, human performance \autocite{openaiGPT4TechnicalReport2023}, but can they simulate the cognitive and psychological characteristics of humans?

If LLMs can indeed simulate human psychological profiles, they could become indispensable to social scientific research. We could run preliminary,
exploratory studies using LLMs for the fraction of a cost. LLMs can also provide another line of evidence towards a specific hypothesis: if the
evidence using human data is of similar strength to what is observed using LLM data, that would increase our confidence in the conclusion. LLM-based
research could also allow us to gather datasets that are difficult or impossible to obtain with humans. Most excitingly, LLMs, consistent with human
psychology, could allow us to create powerful LLM-powered agent-based simulations of different environments, such as social media, a particular
organisation, or entire countries \autocite{gaoS3SocialnetworkSimulation2023,heHomophilyArtificialSocial2023, wangSurveyLargeLanguage2023,
    kwonWHATWHENHOW2023, liLargeLanguageModelEmpowered2023, liMetaAgentsSimulatingInteractions2023,
    parkGenerativeAgentsInteractive2023,wangSurveyLargeLanguage2023}. This could serve as a behavioural sandbox, where various hypotheses, ranging from
the effects different news feed algorithms \autocite{tornbergSimulatingSocialMedia2023}, recommendation systems \autocite{wangSurveyLargeLanguage2023}
or country-wide policies \autocite{gaoS3SocialnetworkSimulation2023}, could be tested before being piloted or implemented.

Social scientists have already begun investigating whether LLMs can simulate
human behaviours by using them as participants to study
cognition\autocite{binzUsingCognitivePsychology2023}, political and public
opinion \autocite{argyleOutOneMany2023,santurkarWhoseOpinionsLanguage2023},
economic decision making
\autocite{hortonLargeLanguageModels2023,parkDiminishedDiversityofthoughtStandard2024},
and theory of mind
\autocite{kosinskiTheoryMindMight2023,sapNeuralTheoryofMindLimits2022,shapiraCleverHansNeural2023,vanduijnTheoryMindLarge2023},
among others \autocite{aherUsingLargeLanguage2023}. To understand their
underlying personality, LLMs have also been used as participants in surveys using
standardized psychological
questionnaires \autocite{deshpandeToxicityChatGPTAnalyzing2023,giorgiSleptBabyUsing2023,guptaInvestigatingApplicabilitySelfAssessment2023,karraEstimatingPersonalityWhiteBox2023,liAreYouMasquerade2023,pellertAIPsychometricsAssessing2022,romeroGPTLanguageModels2023,stevensonPuttingGPT3Creativity2022}.
Many of those studies have explored the ability of LLMs to simulate
human behaviours \autocite{caronIdentifyingManipulatingPersonality2022,huangWhoChatGPTBenchmarking2023,jiangPersonaLLMInvestigatingAbility2023,miottoWhoGPT3Exploration2022}.
For instance, \textcite{huangRevisitingReliabilityPsychological2023} examined
the personality of GPT-3.5 using a standardized personality scale, varying
instructions, item phrasing, language, choice labels, and choiceordering,
resulting in 2,500 configurations. Although they conclude that GPT-3.5 shows
“consistency in
responses” \parencite[1]{huangRevisitingReliabilityPsychological2023}, their
analyses are limited, focused on summary scores, and do little to
psychometrically falsify the conclusion. There is also a comparable number of
studies that use similar methodology and find mixed evidence of LLMs’ ability to
simulate human behaviours \autocite{aherUsingLargeLanguage2023,aiCognitionActionConsistent2024,binzUsingCognitivePsychology2023,huQuantifyingPersonaEffect2024,parkDiminishedDiversityofthoughtStandard2024,santurkarWhoseOpinionsLanguage2023}.

One particularly thorny problem when treating LLMs as survey participants is
that using the same prompt to obtain model responses to a survey item multiple
times results in close-to-uniform response distributions (e.g., see Experiment 1
in \citeauthor{huangRevisitingReliabilityPsychological2023},
\citeyear{huangRevisitingReliabilityPsychological2023};
\citeauthor{parkDiminishedDiversityofthoughtStandard2024},
\citeyear{parkDiminishedDiversityofthoughtStandard2024}), which is of limited
utility for inferential analysis. Modifying model hyper-parameters, such as
temperature, also does not seem to solve the issue
\autocite{argyleOutOneMany2023,serapio-garciaPersonalityTraitsLarge2023}.

\textcite{serapio-garciaPersonalityTraitsLarge2023} proposed a solution to this
problem by encouraging variation in item responses using generic persona
prompting: ask the LLM to adopt a generic persona description and respond to the
survey items as that persona. The authors incorporated randomly sampled
descriptions from the PersonaChat dataset
\autocite{zhangPersonalizingDialogueAgents2018} into multiple choice question
answering tasks and found that Google’s PaLM models provided reliable and valid
personality test responses. The idea behind generic persona prompting is to
ensure that the LLM is not influenced in any particular direction, and thus, its
responses represent a more random sample of the corpus on which it was trained.

Another solution of greater interest is to use more targeted persona
descriptions, an approach referred to as silicon persona prompting.
\textcite{argyleOutOneMany2023}, introduced this approach in the context of
eliciting voting preferences from LLMs: they found that when GPT-3.5 was
prompted using persona descriptions, created from basic demographic variables
(age, sex, ethnicity, religious attendance, etc.), it was able to reproduce
human voting preferences. Crucially, the authors used census data, so they were
able to calculate differences in voting preference distributions between actual
humans from the census data and LLMs that assumed equivalent personas.

Regardless of what kind of persona descriptions are used in the prompting,
however, the problem of understanding the psychology of LLMs from text responses
is challenging, rendering the extant conflicting findings unsurprising. To
address this issue, we use frameworks from psychometrics, the science of
psychological measurement \autocite{rustModernPsychometricsScience2021}.
Psychometrics offer a robust and rigorous methodology for capturing underlying,
or latent, constructs in generative AI systems that give rise to behaviours
across a range of tasks
\autocite{pellertAIPsychometricsAssessing2022,wangWhenLargeLanguage2023}.

Our aim was to conduct a comprehensive, psychometrically-informed analysis of
LLM abilities to simulate human responses both using generic persona prompting
and silicon persona prompting. Broadly speaking, with generic persona prompting,
we expected LLM responses to be similar to what would be expected from a sample
of the population, while with silicon persona prompting, we expected that the
responses would closely match those of the targeted personas. To do this, we
used OpenAI’s flagship models, GPT-3.5 and GPT-4, due to their growing
popularity and state-of-the-art performance
\autocite{openaiGPT4TechnicalReport2023}.

To construct our silicon samples, we used a representative human sample from a
large survey conducted in the UK in collaboration with the British Broadcasting
Corporation (BBC) between November 2009 and April 2011 \footnote{All files are
    available from the UK Data Archive database (SN 7656—BBC Big Personality Test,
    2009–2011: Dataset for Mapping Personality across Great Britain). The catalogue
    record created for this data collection can be viewed at the following URL:
    \url{http://discover.ukdataservice.ac.uk/catalogue/?sn=7656}.}
\autocite{rentfrow_regional_2015}. The survey received 588,014 responses and
covered various questions, including demographics, education, work,
relationships, personality, and health. For the current study, we focus on a
subset of variables, including demographics and personality scores. After
excluding entries with any missing responses and duplicates, we use a resulting
subset of N=123,828 responses as our ground truth human dataset to compare with
LLM responses.

\section{Methods}
\label{sec:methods}

Figure \ref{fig:method} summarises our methodology. Data and code are available
at \url{https://github.com/nikbpetrov/LLMs-Simulate-Humans}.

We created a structured prompting template, similar to that used by
\textcite{serapio-garciaPersonalityTraitsLarge2023}. The template included a
personality instruction, persona description, test instruction, item statement,
and item postamble. Given that
\textcite{serapio-garciaPersonalityTraitsLarge2023} found that test instruction
and item postamble variations did not yield substantial differences in model
responses, we did not vary those. We then used this template to query GPT-3.5
and GPT-4.

We sent two sets of queries to each model, with each set using a different
persona description.

For the first set of queries, we used 150 generic persona descriptions randomly
sampled from the PersonaChat dataset
\autocite{zhangPersonalizingDialogueAgents2018}. Each description was four or
five sentences long.

For the second set of queries, we used 1,000 silicon personas
\autocite{argyleOutOneMany2023} based on the BBC personality dataset. After
minimal data processing of the original dataset (selecting only needed
variables, removing blank responses and duplicates), we were left with
N=123,828, of which we selected 1,000 entries randomly. To construct persona
descriptions, we used the following variables: age, sex, ethnic background,
country, education (self and parents’), occupation, income, relationship status,
and number of children (see Appendix \ref{app:variable_transformations_bbc} for details).

As part of the template, we partially replicated the design of
\textcite{serapio-garciaPersonalityTraitsLarge2023}. We administered a
personality test, namely the Big Five Inventory , alongside eight tests of
personality-related constructs: the Positive Affect and Negative Affect
subscales of the Positive and Negative Affect Schedule (PANAS;
\citeauthor{watsonDevelopmentValidationBrief1988},
\citeyear{watsonDevelopmentValidationBrief1988}), the Buss-Perry Aggression
Questionnaire’s (BPAQ) Physical Aggression, Verbal Aggression, Anger, and
Hostility subscales \autocite{bussAggressionQuestionnaire1992}, and the Creative
Self-Efficacy and Creative Personal Identity subscales of the Short Scale of
Creative Self (SSCS; \citeauthor{karwowskiItDoesnHurt2011},
\citeyear{karwowskiItDoesnHurt2011}), resulting in a total of 104 items. We
varied the item statement and item postamble based on the test (see Appendix \ref{app:psychological_scales}
for details on these measures).

\begin{figure*}
    \centering
    \includegraphics
    [trim=0 0 0 0,clip, width=0.95\textwidth]
    {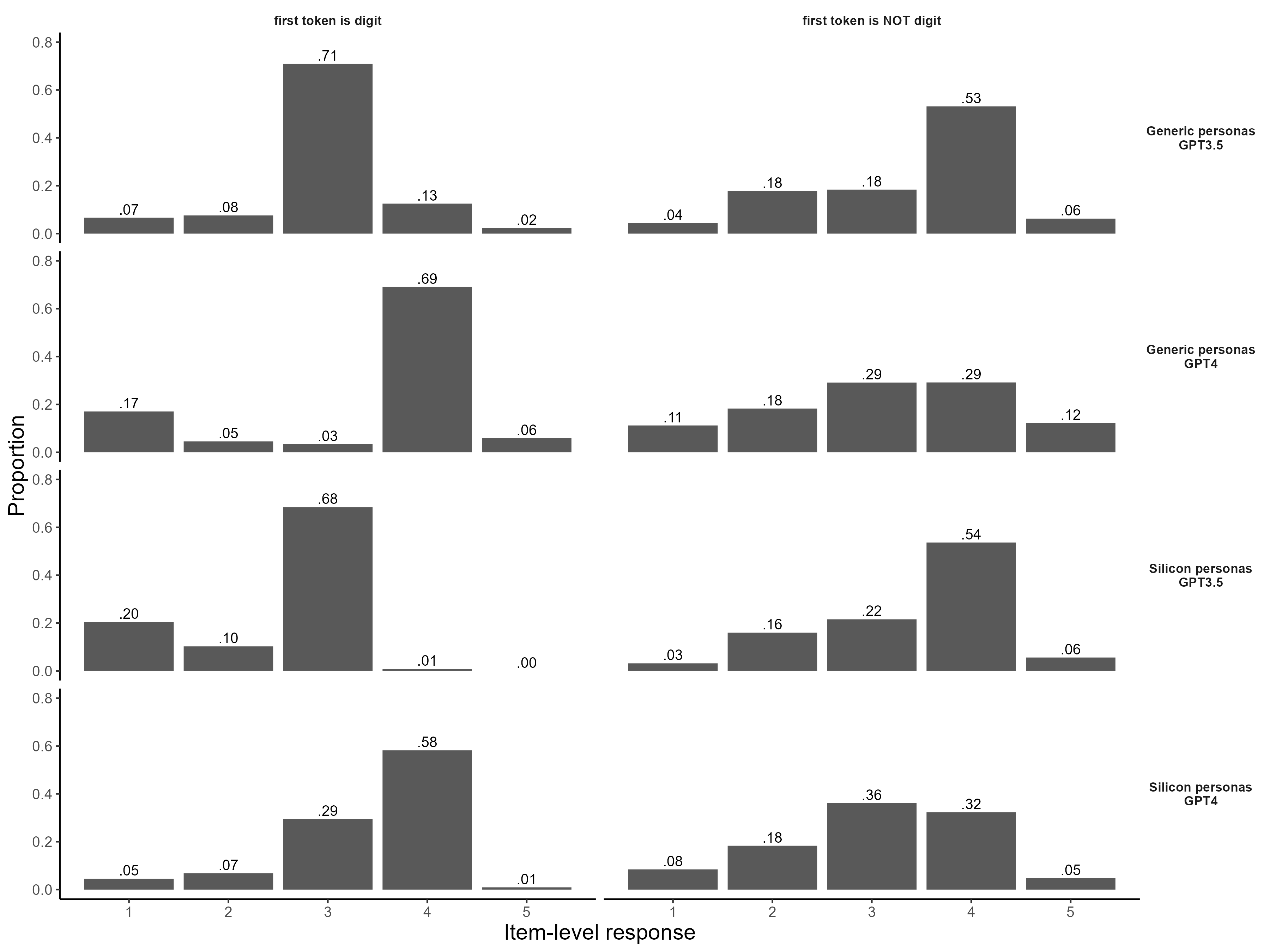}
    \caption{
        \small\textbf{Item-level response distributions across all responses split by whether the first token was a digit or not} LLMs (rows) produce text responses which we further processed to extract a numeric response. Past research has used only responses whose first digit is a numeric one and discarded the rest. Here, we compare the relative frequency (y-axis) of item-level responses across all survey items (x-axis), split by whether the first token was a digit (columns). We see very different distributions of responses across all LLMs and prompting variations when the first token is a digit one vs a non-digit one.
    }
    \label{fig:first_token_digit}
\end{figure*}

Using this template, we constructed a total of 239,200 prompts. Each prompt was
part of a separate OpenAI API\footnote{\url{https://platform.openai.com}}
call\footnote{An alternative approach would be to include “response memory”,
    such that the questions the LLM has access to its previous responses. However,
    \textcite{miottoWhoGPT3Exploration2022} found that this does not make a
    statistically significant difference in aggregate scores when they tested
    GPT-3.5 with a human values scale.} to ‘gpt-3.5-turbo-1106’ and
‘gpt-4-1106-preview’ using a temperature of 0. Unlike
\textcite{serapio-garciaPersonalityTraitsLarge2023}, we did not have access to
the log probabilities of possible model completions to prompts, as OpenAI’s API
does not allow access to these at the time of running this experiment (January
’24). To circumvent this problem, some have restricted the maximum tokens of the
GPT models’ responses to 1 and discarded non-digit responses
\autocite{binzUsingCognitivePsychology2023,dornerPersonalityTestsGeneralize2023}.
In our testing, we found that if we allowed the model to respond freely, while
sometimes the first token was a non-digit one, usually the sentences still
contained a sensible response. We also observed that when we were looking only
for a single-digit response, or using only responses where the first character
was a digit, the resulting response distribution was close to uniform with an
overwhelming proportion of “3”s (after we collected the full dataset, we
confirmed this initial observation – see Figure \ref{fig:first_token_digit}). To
incorporate this insight, we restricted the maximum tokens of the responses to
50 for GPT-3.5 and to 200 for GPT-4 (during testing we found that GPT-4 took
more tokens to generate a numeric answer). For each text response we took the
first digit (regardless of its position) and considered this an item response.

Overall, we executed 239,200 API calls: 15,600 were made using generic persona
descriptions to both GPT-3.5 and GPT-4, and 104,000 were made using silicon
samples to both GPT-3.5 and GPT-4.

\section{Results}
\label{sec:results}

We conducted a psychometrically comprehensive analysis of our tested LLMs’
responses across several metrics. First, we examined the reliability (internal
consistency) of the responses (i.e. do the responses form consistent patterns
across different items). Next, we studied the construct validity of their
responses (i.e., whether the responses actually reflect the underlying
construct) in the form of internal validity (intercorrelations between Big Five
traits) and criterion validity in the form of correlations between different
constructs (e.g., Extraversion correlated with Positive and Negative affect
scores). As a further validity examination of the LLM responses, we looked at
the proportional frequency of item-level responses, as this would help us
understand whether the summary scores are reached in the same way as with
humans. As a yet further validity test, we checked the structural validity of
LLMs’ responses by fitting a confirmatory factor analysis model, expecting a
similarly good fit as with human data both in terms of path loadings as well as
fit indices. Last but not least, for the silicon personas, we tested how
accurately the personality scores based on LLMs’ data match those of the actual
humans, when conditioning on the same background information. Across all our
analyses, we break the results down by persona type (generic and silicon) and
LLM (GPT-3.5 and GPT-4). Where relevant, we include data from humans for
comparison.

First, in order to examine the internal consistency of all subscales, we used
three metrics: Cronbach’s $\alpha$
\autocite{cronbachCoefficientAlphaInternal1951}, Greatest Lower Bound
(\textit{glb}; \citeauthor{woodhouseLowerBoundsReliability1977},
\citeyear{woodhouseLowerBoundsReliability1977}), and McDonald’s $\omega$
\autocite{mcdonaldTestTheoryUnified1999}. The combination of these metrics
allowed us to paint a comprehensive picture of the performance of the LLMs
across the subscales. While there were few differences between those
coefficients, overall, they showed a similar pattern when comparing across
models and subscales
\autocite{hayesUseOmegaRather2020,sijtsmaUseMisuseVery2009,sijtsmaPartIIUse2021,taberUseCronbachAlpha2018,trizano-hermosillaBestAlternativesCronbach2016,vaskeRethinkingInternalConsistency2017}.
Thus, hereon, we focus our attention on $\alpha$ as it is the most widely used
measure and it consistently showed the lowest value, giving us a lower bound of
internal consistency.

\begin{figure*}
    \centering
    \includegraphics
    [trim=0 0 0 0,clip, width=0.95\textwidth]
    {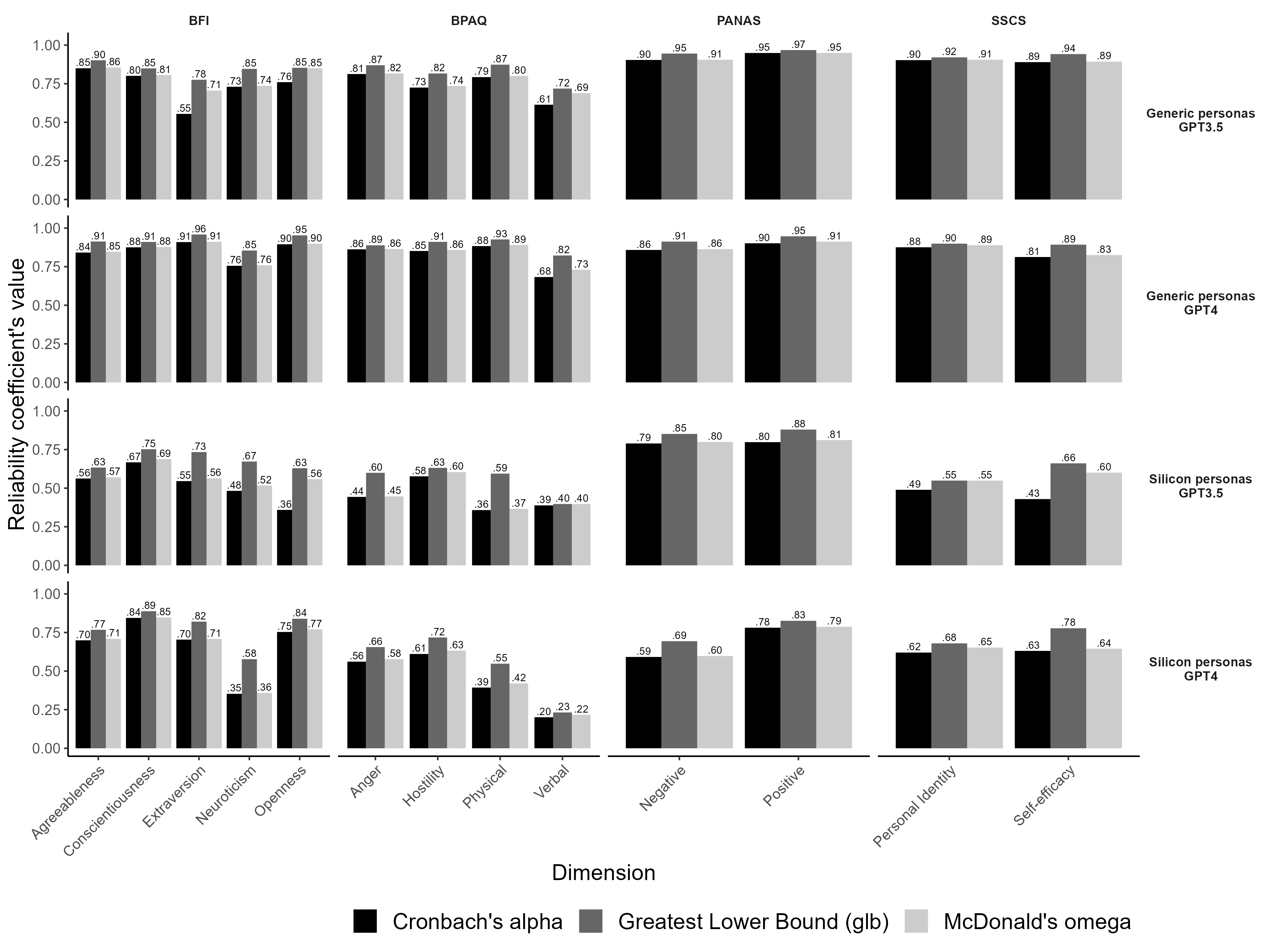}
    \caption{
        \small\textbf{Internal consistency of all measures across LLM models and prompting styles} We computed three reliability indices (colour) for every questionnaire (columns) and its subscales (x-axis) across all LLMs and prompting variations (rows). The plot shows that a) differences between reliability indices are few and b) the reliability of the data from LLMs, when using silicon sampling, can be very low <.70 for some measures.
    }
    \label{fig:reliability}
\end{figure*}
For Cronbach’s $\alpha$ values, we expected the values to exceed the commonly
used threshold of .70 used in psychometrics
\autocite{nunnallyPsychometricTheory1978}. As can be seen from Figure \ref{fig:reliability}, when
generic personas were used with either GPT-3.5 and GPT-4, the data exhibited
acceptable internal consistency: all subscale $\alpha$ values were $\ge$ .70, with
the exception of those for Extraversion and Verbal Aggression, where GPT-3.5
fared worse (.58 and .63, respectively). Reliability was significantly worse for
the silicon personas with GPT-3.5 and GPT-4, where $\alpha$ values were $\ge$ .70
for only 3 or 4 of the 13 subscales, respectively. Notably, there were some
values in the .10–.50 range (6 out of 13 for GPT-3.5 and 3 out of 13 for GPT-4).
This pattern of findings calls into question the test-level reliability of LLMs’
responses using silicon personas.

\begin{figure*}
    \centering
    \includegraphics
    [trim=0 0 0 0,clip, width=0.95\textwidth]
    {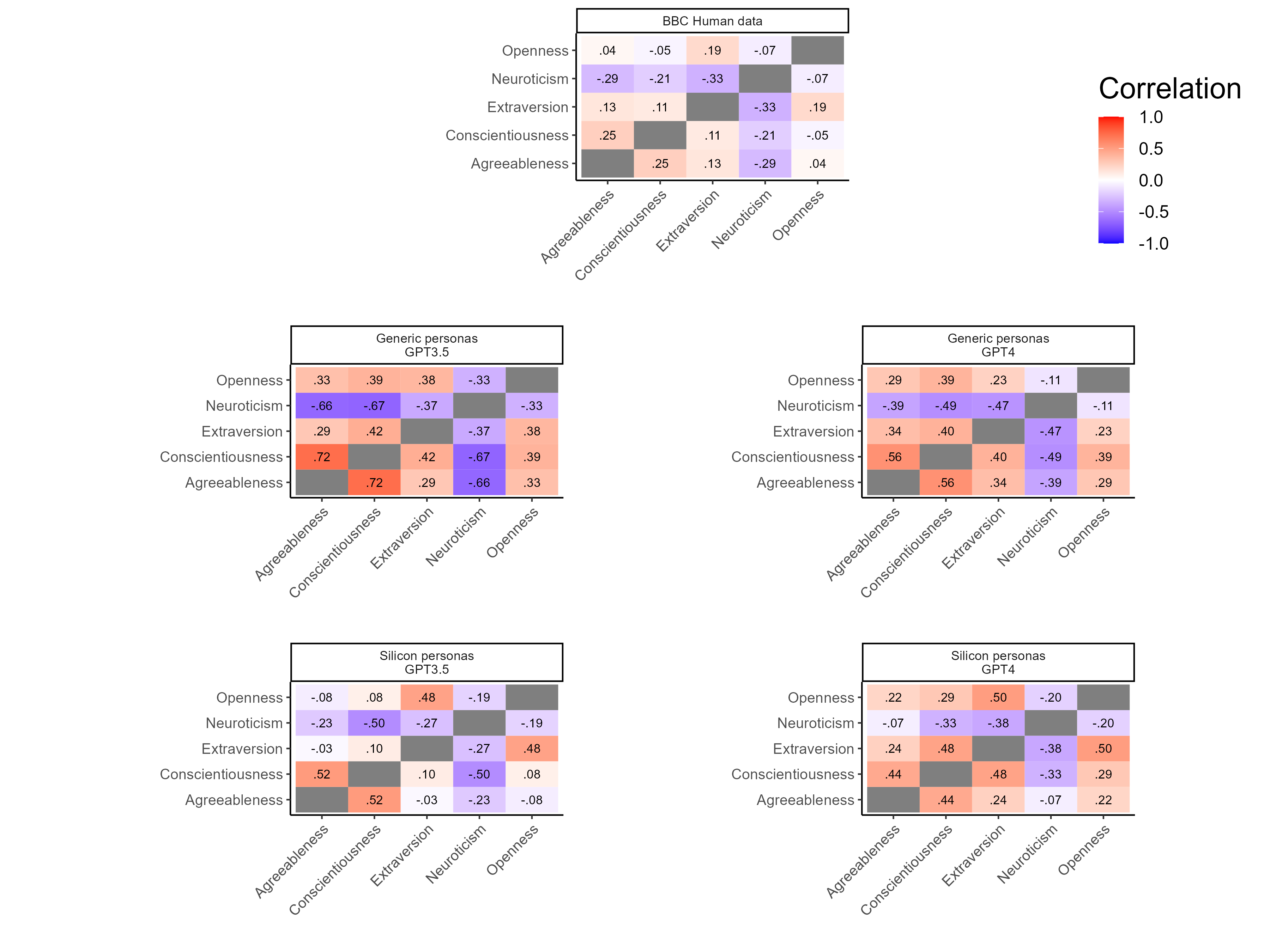}
    \caption{
        \small\textbf{Intercorrelations between Big Five traits across LLM models and prompting styles} Intercorrelations between Big Five traits were computed for the tested LLMs across prompting styles (bottom two rows) and human data from a large representative sample \autocite{rentfrow_regional_2015} is shown on the top row. The plots show that data from LLMs tend to produce much higher intercorrelations.
    }
    \label{fig:intercorrelations}
\end{figure*}
Next, we examined the internal validity of the LLMs’ responses by computing the
intercorrelations between Big Five traits and comparing them to those observed
in human data. As can be seen from Figure \ref{fig:intercorrelations}, when the generic personas were used
with either GPT-3.5 or GPT-4, the resultant correlations were substantially
higher relative to those in human data. When silicon personas were used with
either LLM, the correlations were still higher than in human data, but the
results were more ambiguous.

\begin{figure*}
    \centering
    \includegraphics
    [trim=0 0 0 0,clip, width=0.95\textwidth]
    {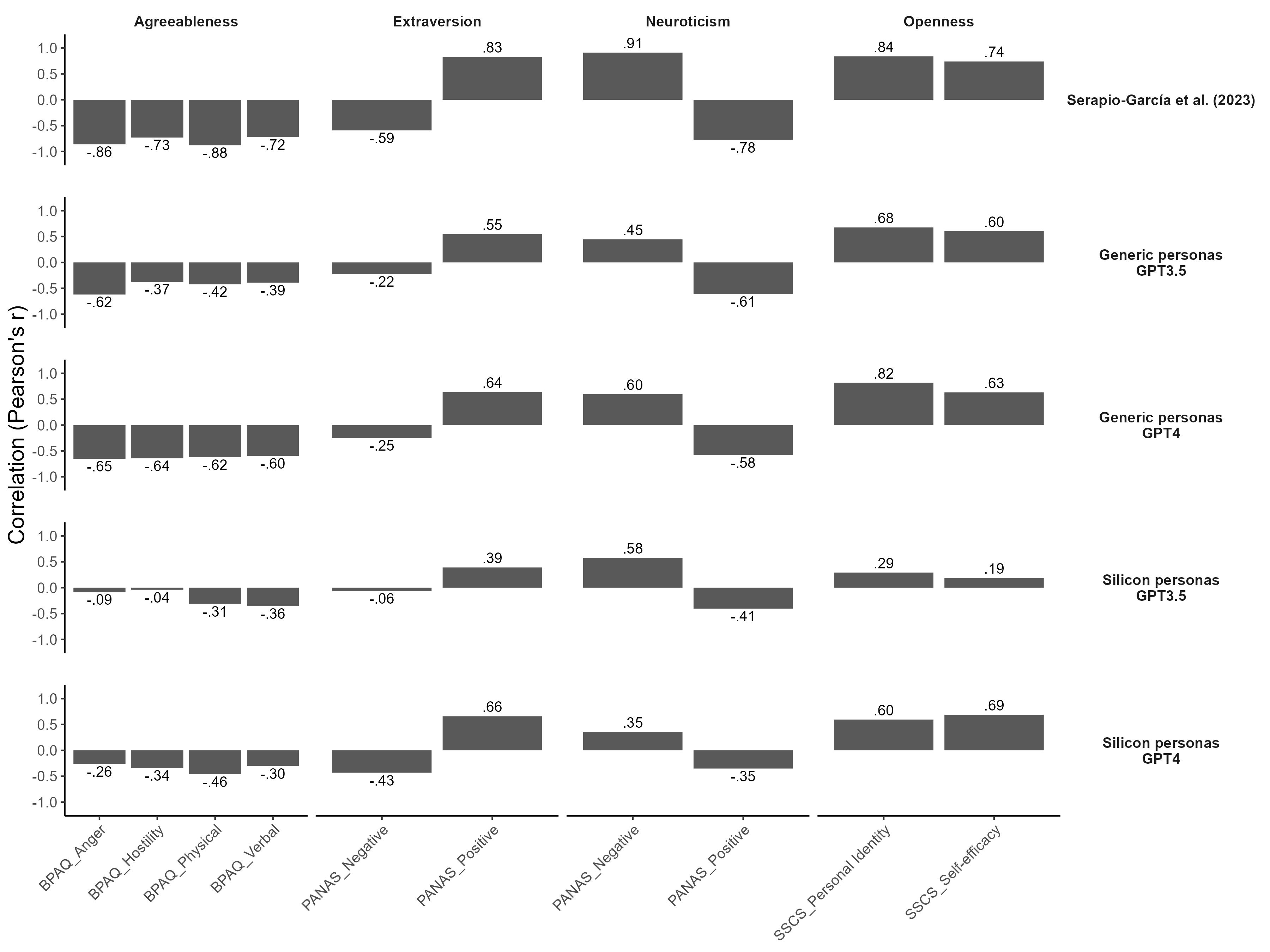}
    \caption{
        \small\textbf{Criterion validity correlations across LLMs and prompting styles} Selected Pearson’s correlations were computed between Big Five traits (columns) and personality-related constructs (x-axis) to test the criterion validity of the tested LLMs (bottom 4 rows). Comparison data from \textcite{serapio-garciaPersonalityTraitsLarge2023} is shown on the top row. NB: \textcite{serapio-garciaPersonalityTraitsLarge2023} used the IPIP-NEO to measure Big Five traits, while we used the BFI, though the authors also show that the two subscales are correlated at >.90.
    }
    \label{fig:validity}
\end{figure*}
We then examined the criterion validity of LLMs’ responses by computing the
correlations between various personality constructs, similar to
\autocite{serapio-garciaPersonalityTraitsLarge2023} and expected similar
results. As can be seen from Figure \ref{fig:validity}, LLM personality test data derived from
generic personas adequately correlated with personality-related criteria tests,
with GPT-4 data faring marginally better. Data generated from silicon personas,
however, showed significantly poorer psychometric properties: with some minor
exceptions, the data from either GPT-3.5 or GPT-4 resulted in weaker
correlations relative to those derived from generic personas and much weaker
relative to what \autocite{serapio-garciaPersonalityTraitsLarge2023} reported.
While this could be due to the poor internal consistency of the tests, it
nonetheless highlights problems with the silicon persona prompting approach.

\begin{figure*}
    \centering
    \includegraphics
    [trim=0 0 0 0,clip, width=0.95\textwidth]
    {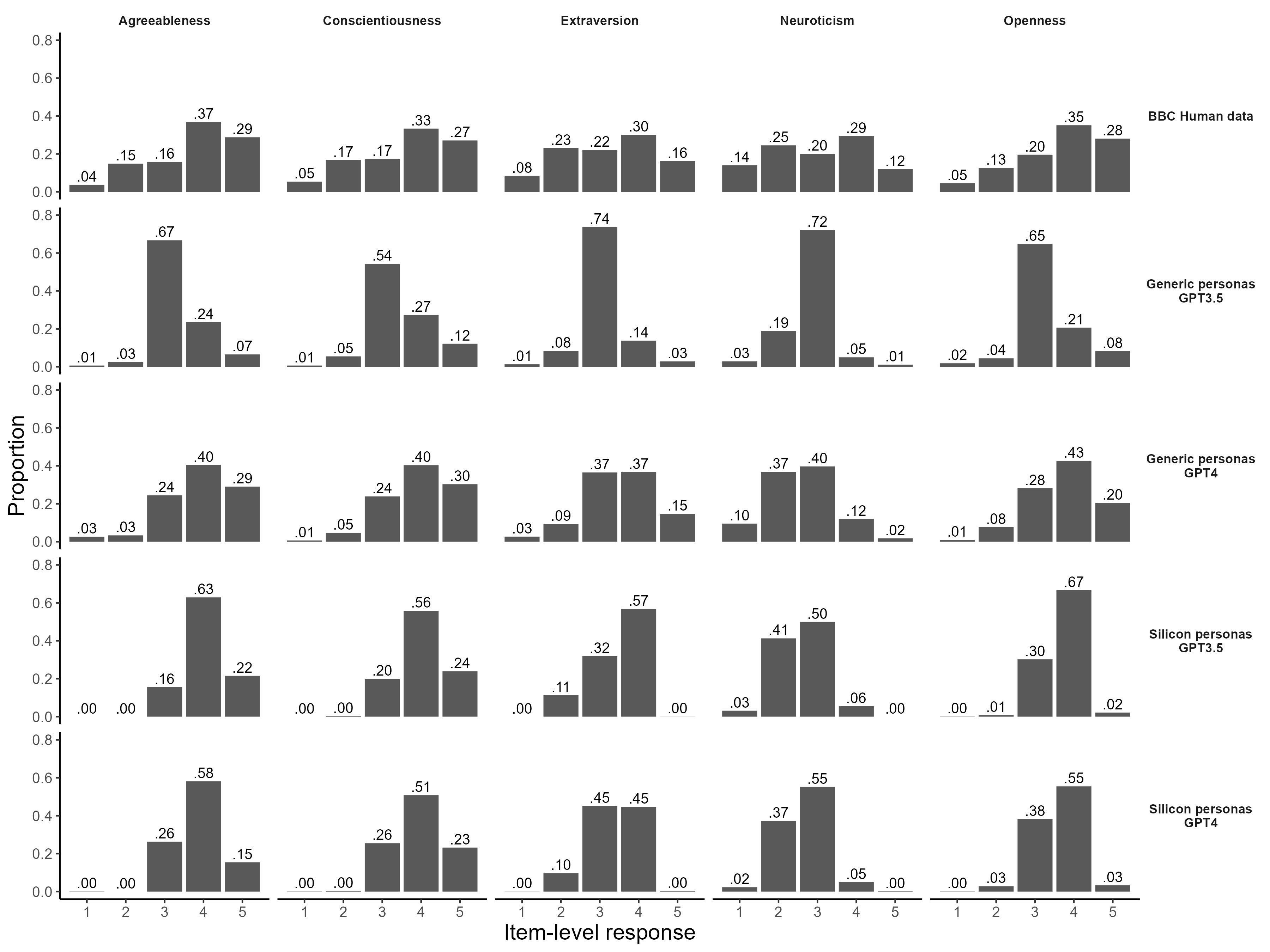}
    \caption{
        \small\textbf{Item-level frequencies across LLMs and prompting styles} The relative frequency (y axis) of item-level responses (x axis) across Big Five traits (columns) and LLMs (bottom 4 rows) is shown, along with the data from a representative human sample (\citeauthor{rentfrow_regional_2015}, \citeyear{rentfrow_regional_2015}; top row). We see that there are differences in the distributions between prompt variations – generic vs. silicon – and that the distributions when using silicon prompting across both GPT-3.5 and GPT-4 are quite different from human ones.
    }
    \label{fig:frequencies}
\end{figure*}
Next, we examined the distributions of raw responses of LLMs to BFI items –
Figure \ref{fig:frequencies}. With generic personas, data from GPT-3.5 fared the worst by producing
close to uniform distributions, with the majority of the responses being “3”.
With silicon personas, data from either GPT-3.5 or GPT-4 produced skewed
distributions in the direction of a ‘desirable’ BFI trait: high Agreeableness,
high Conscientiousness, high Extraversion, low Neuroticism and high Openness.
While these were not uniform, they were clustered around responses “2” and “4”.
The data from GPT-4 with generic personas showed promising distributions, most
closely matching those in human data.

\begin{figure*}
    \centering
    \includegraphics
    [trim=0 0 0 0,clip, width=0.95\textwidth]
    {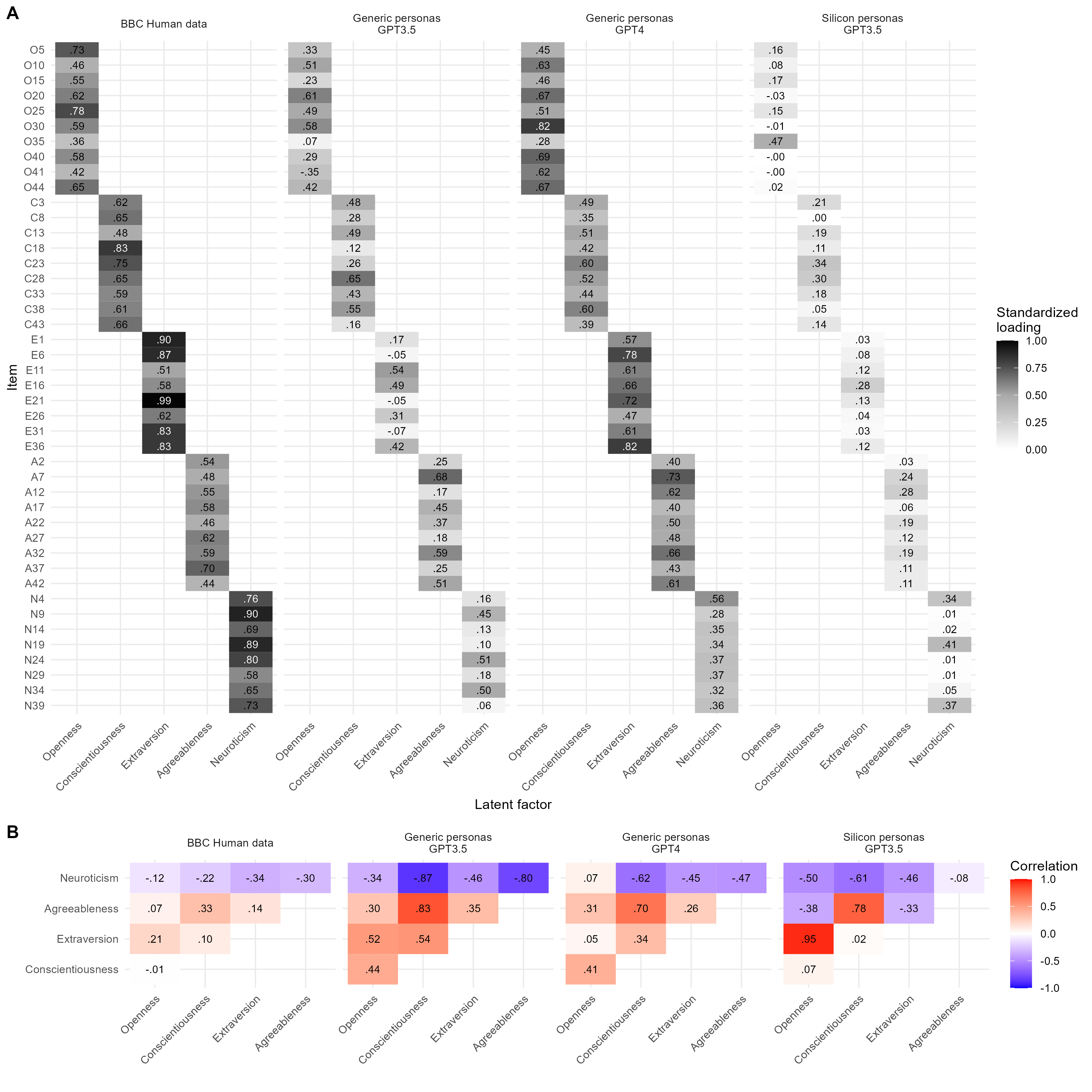}
    \caption{
        \small\textbf{Results from confirmatory factor analyses on the BFI scale across all models and prompting styles} Summary of the results from confirmatory factor analyses with the same model being fit on different data sources (columns). Panel A shows the path loadings from each observed variable (i.e., item, y-axis) to its latent factor (i.e., dimension, x-axis). Panel B shows the covariances between latent factors. Overall, relative to human data, data from LLMs shows smaller path loadings and higher latent factor covariances.
    }
    \label{fig:cfa_results_bfi}
\end{figure*}
We then extended our examination of the BFI items by checking their structural
validity. We ran confirmatory factor analyses on the responses of different
models expecting to recover the Big Five structure as with human data (we did
the same for the rest of the scales and reached similar conclusions as with the
BFI, see Appendix \ref{app:cfa_bpaq_sscs_panas}). For each data source, we performed a confirmatory factor
analysis using a maximum-likelihood estimation with robust standard errors;
results are visualised in Figure \ref{fig:cfa_results_bfi}. For each model, we report three fit
indices\footnote{Notably, the fit indices, even for the ground truth human data,
    seem lower than acceptable thresholds \autocite{hairMultivariateDataAnalysis2019}, though
    we reserve any judgement as to why that might be. Extant literature points to
    differences between confirmatory factor analysis models and exploratory
    structural equation models
    \autocite{gomesComparingESEMCFA2017,marshNewLookBig2010,marshExploratoryStructuralEquation2014}.
    In any case, we focus our interpretations on relative differences between
    models, not absolute interpretations of the models’ fit.}: one absolute (GFI),
one relative (IFI), and one based on the noncentrality parameter (RMSEA)
\autocite{byrneStructuralEquationModeling2010,newsomClarificationsRecommendationsFit2018} – see Table \ref{tab:cfa_fit_bfi}.
Notably, the model estimation for the data from GPT-4 using silicon personas
failed as the covariance matrix was not positive definite (one eigenvalue was
negative), most likely due to high covariances between latent variables,
suggesting that the Big Five factor structure was not recoverable from the data.
As for the remaining models, for both GPT-3.5 with generic personas and GPT-3.5
with silicon personas, the loadings were all quite low (majority of loadings
<.50), while for GPT-4 with generic personas, the loadings were comparable to
those from human data, with the exception of the Neuroticism factor, where all
loadings, except one, were below .40. Examining the fit indices (Table \ref{tab:cfa_fit_bfi}), we
witness a classic case of a reliability paradox
\autocite{hancockReliabilityParadoxAssessing2011,mcneishThornyRelationMeasurement2018}
for the data from GPT-3.5 (both generic and silicon personas): despite poor data
quality and thus low path loadings, fit indices indicate a decent fit,
comparable to that from human data. The IFI, our relative fit index, is less
susceptible to this paradox \autocite{milesTimePlaceIncremental2007}, which is
why it shows a poorer fit compared to our other indices. The data from GPT-4
with generic personas performs worse on the fit indices, relative to human data,
most likely due to the weakly-performing Neuroticism factor. Overall, we
conclude that data from current-generation OpenAI LLMs conditioned on our two
prompting styles do not recover the expected factor structure and thus have do
not show sufficient structural validity.

\begin{table}[ht]
    \centering
    \caption{
        \small \textbf{Fit indices from the confirmatory factor analyses on the BFI scale across different data sources.}
        A confirmatory factor analysis using a maximum-likelihood estimation with robust standard errors was performed for each data source and 3 fit indices reported.
    }
    \label{tab:cfa_fit_bfi}
    \begin{tabular}{@{}p{0.45\columnwidth}ccc@{}}
        \toprule
        Data source               & GFI & IFI & RMSEA \\ \midrule
        Human data (BBC)          & .81 & .75 & .06   \\
        Generic sampling, GPT-3.5 & .62 & .65 & .09   \\
        Generic sampling, GPT-4   & .52 & .64 & .11   \\
        Silicon sampling, GPT-3.5 & .80 & .57 & .06   \\ \bottomrule
    \end{tabular}
\end{table}
Last but not least, for the responses that used silicon personas, we examined
how well the LLMs replicated the responses of humans given the same persona
description. To do that, for each persona, we took the absolute difference
between the trait score from the LLMs’ data and that of the actual person, i.e.
the ground truth trait score. We call this measure ‘trait bias’, with higher
scores indicating higher mismatch between the trait that the LLM data produces
and that of the human with the same background information.

\begin{figure*}
    \centering
    \includegraphics
    [trim=0 0 0 0,clip, width=0.95\textwidth]
    {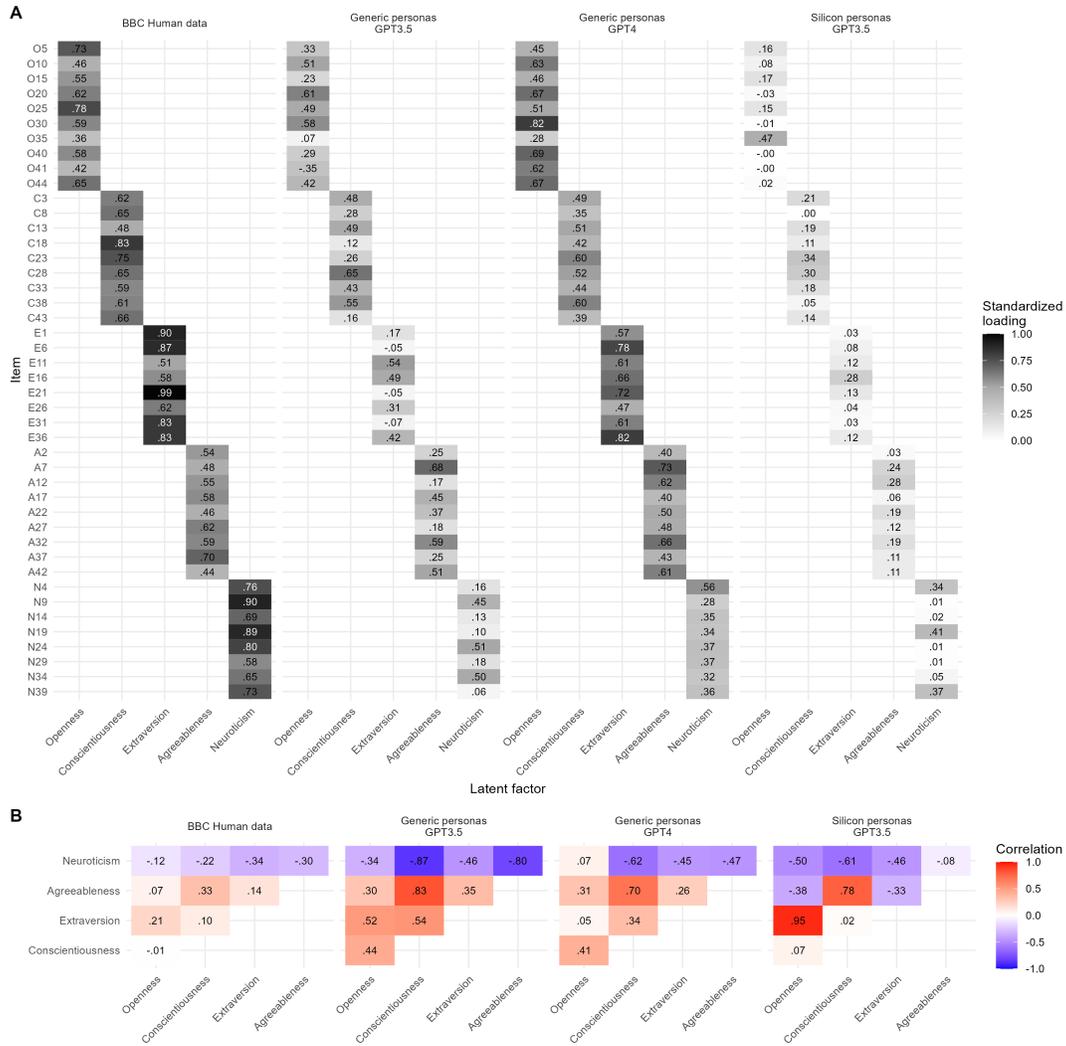}
    \caption{
        \small\textbf{Bias of LLM-assumed traits when using silicon personas} We calculated the absolute difference between personality scores on each trait for human and LLM-assumed personality, conditioned on the same demographic information – this difference, which we call trait bias, is on the y-axis. Across traits, we find no statistically significant differences between GPT-3.5 and GPT-4, except for BFI Agreeableness, with average trait bias across all traits at .63 and .62 for GPT-3.5 and GPT-4, respectively.
    }
    \label{fig:trait_bias}
\end{figure*}
We found a statistically significant difference between GPT-3.5 and GPT-4 in
their trait bias regarding BFI Agreeableness, \textit{t}(1968.5) = 2.83,
\textit{p} = .005, but not for any other traits (all other \textit{p}s > .647),
with the average bias across all traits being \textit{M}=.63 (\textit{SD}=.25)
for GPT-3.5 and \textit{M}=.62 (\textit{SD}=.24) for GPT-4 – see Figure \ref{fig:trait_bias}.

\begin{figure*}
    \centering
    \includegraphics
    [trim=0 0 0 0,clip, width=0.95\textwidth]
    {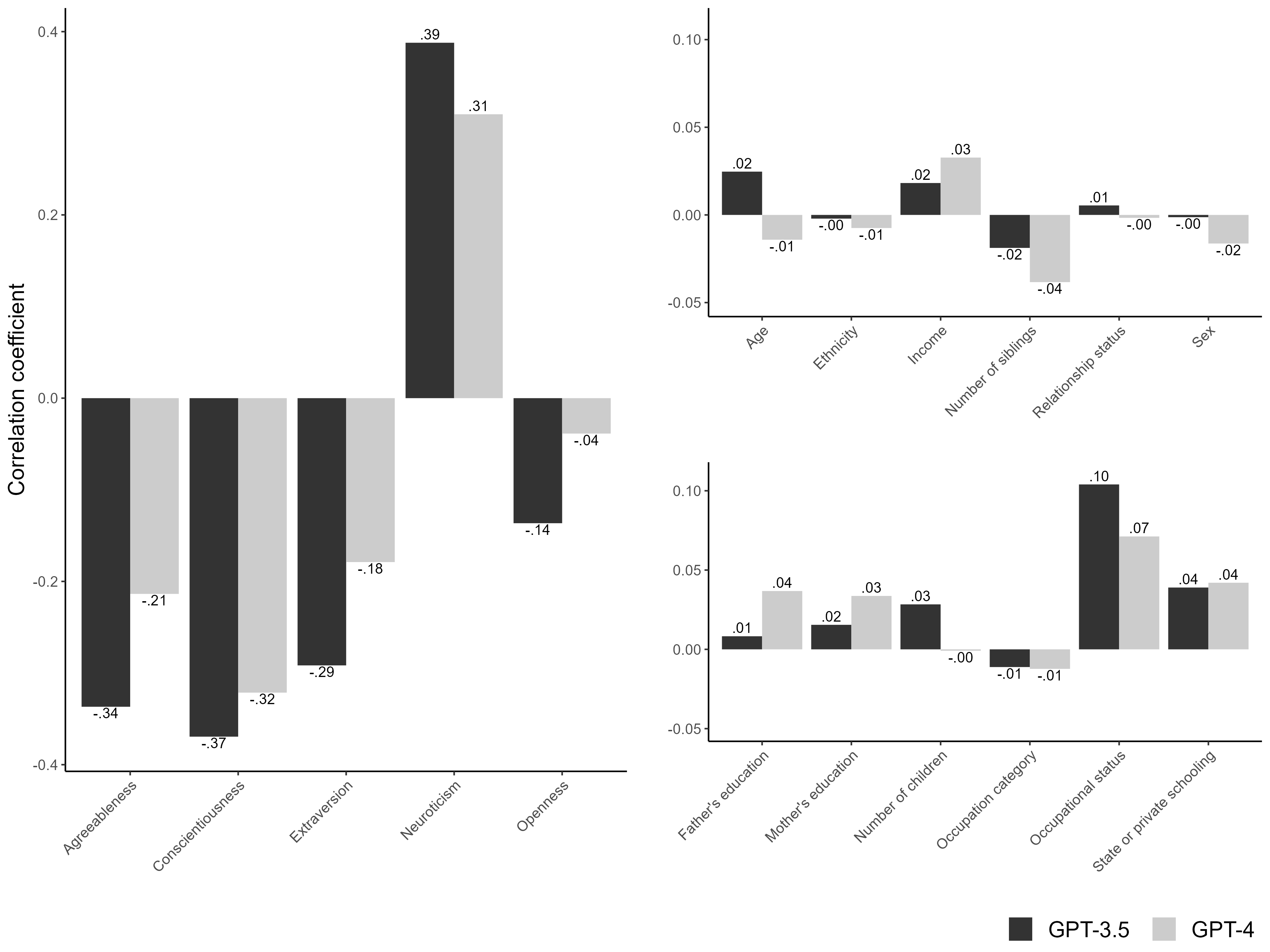}
    \caption{
        \small\textbf{Correlations between ground truth demographic variables of silicon samples and average trait bias of LLM models} We correlated the average trait bias (i.e. the absolute difference between personality scores of humans and LLMs, conditioned on the same background information) across all traits with the demographic variables (right-hand side) and the personality scores of humans (left-hand side). The plots show that the inaccuracy of personality scores displayed by LLMs (i.e. high trait bias) is not correlated with demographic variables but is positively correlated with Neuroticism, and negatively correlated with the other Big Five traits, suggesting that LLMs are less capable of representing individuals who tend to be more disagreeable, unconscientious, introverted, emotionally stable, and not open to experience.
    }
    \label{fig:trait_bias_corrs}
\end{figure*}
We further examined the source of the trait bias by correlating the average bias
across all traits with all variables from the human data. We found that the
trait bias did not correlate with any of the demographic variables, but it
showed moderate correlations with personality traits. Specifically, we found
that greater failure of the LLMs to simulate human behaviour (i.e. high trait
bias) was associated with lower agreeableness, lower conscientiousness, lower
extraversion, higher neuroticism and lower openness, with GPT-4 consistently
showing marginally smaller correlations relative to GPT-3.5 – see Figure \ref{fig:trait_bias_corrs}.

\section{Discussion}
\label{sec:discussion}

Whether it’s reducing costs, collecting hard-to-acquire data, or conducting
full-scale simulations using simulated agents, the value of using LLMs as a tool
that can simulate human behaviour is undeniable. With an initial wave of
investigations showing conflicting evidence, a robust and rigorous approach is
needed to map out the psychosocial abilities of LLMs.

To this end, we conducted a psychometric investigation of whether OpenAI’s
flagship LLM models, namely GPT-3.5 and GPT-4, can be prompted to simulate human
psychological behaviours. We asked both models to adopt two different kinds of
personas: either generic ones, described using four or five short random
sentences, or silicon ones, described using mostly demographic variables of
actual humans from a large representative survey.

Overall, our results show that GPT-4, but not GPT-3.5, can be prompted to
simulate generic personas, similar to Google’s PaLM models
\autocite{serapio-garciaPersonalityTraitsLarge2023}. However, our additional
psychometric analyses add some caveats to these results: the data from GPT-4
still showed higher intercorrelations between the Big Five traits compared to
what is seen in human data and its responses did not show solid structural
validity as we were unable to fully recover the Big Five latent factors.

More discouragingly, however, we found that both GPT-3.5 and GPT-4 failed in
simulating human responses using silicon personas. Specifically, when the LLMs
were prompted to assume a specific persona, their responses showed poor
psychometric properties across the board. This is in contrast to
\textcite{argyleOutOneMany2023}, whose silicon persona approach we used. They
found that LLMs simulated political opinions based on demographic data. We think
the likely reason for this is that they aimed to simulate surface-level
correlations between demographic information and voting preferences. In
contrast, we examined latent traits. As such, it is conceivable that information
(i.e., word co-occurrences) about demographics and voting preferences are more
strongly present in the LLMs’ training data compared to latent constructs,
which, by definition, require deeper-level representations.

We note two additional findings from our study. First, across the board, data
from GPT-4 exhibited better psychometric properties than data from GPT-3.5. This
may reflect the overall better performance of GPT-4 compared to GPT-3.5 across
various LLM benchmarks \autocite{openaiGPT4TechnicalReport2023}. This converges
with the data from \textcite{serapio-garciaPersonalityTraitsLarge2023}, who
found that larger PaLM models generated data with better psychometric
properties. We do note, however, that we remain agnostic as to what exactly
drives GPT-4 to better represent latent traits in our study (not least of all
because of the fact that both models are closed-source). It could be because of
a higher number of parameters, different architectures, more (or better quality)
training data, differences in fine-tuning, or a combination of all of the above.
We encourage future research to investigate the effect of different training
variations on the ability of an LLM to represent latent traits. Second, we found
evidence that instruction fine-tuning drives LLMs to behave in a more desirable
way. Using our trait-bias metric, we found that LLMs were more accurate in
reproducing the personality profile of a person who is more agreeable, more
conscientious, less neurotic, more open to experience, and more extraverted. We
speculate this is, at least in part, due to the instruction fine-tuning by
OpenAI to align LLMs to behave in a more congenial, amicable way.

Our study is not without its limitations. Most notably, our examination, while
comprehensive in its analysis, was limited in its method: we relied on a
mono-method approach, only using self-report data. While we believe that this
type of data is the most useful in order to examine the ability of LLMs to
represent latent traits, we think that gathering other kinds of data, such as
asking LLMs to respond to scenarios \autocite{hagendorffDeceptionAbilitiesEmerged2023,kangValuesOpinionsPredicting2023},
multi-turn conversation, or asking it to produce text \autocite{jiangPersonaLLMInvestigatingAbility2023,serapio-garciaPersonalityTraitsLarge2023}, could enrich our understanding
\autocite{campbellConvergentDiscriminantValidation1959}. Furthermore, our attempt to prompt LLMs to replicate the performance of a
specific subpopulation relied on silicon persona prompting, which relied on
individual demographic data. While we think LLMs should ultimately be able to
indirectly extrapolate personality profiles on the basis of demographic
variables, as there are known correlations between personality and demographics
\autocite{goldbergDemographicVariablesPersonality1998,sotoAgeDifferencesPersonality2011}, these associations are not strong.
This could partly explain why LLMs were better at making consistent inferences
using the generic personas: personality is defined as characteristic patterns of
thinking, feeling, and behaving and the generic personas contained more specific
references to such patterns. Future research can use human datasets that contain
ground truth personality scores, in addition to both demographic variables as
well as other data on characteristic patterns of thinking, feeling and
behaviour, such as hobbies or interests, and then use those as seeds to create
silicon personas.

Overall, our results, while contradicting earlier findings that suggested the
ability of LLMs to simulate human behaviours
\autocite{caronIdentifyingManipulatingPersonality2022,huangRevisitingReliabilityPsychological2023,huangWhoChatGPTBenchmarking2023,jiangPersonaLLMInvestigatingAbility2023,miottoWhoGPT3Exploration2022},
are in line with recent findings that cast doubt on current LLM abilities.
Specifically, \textcite{aiCognitionActionConsistent2024} Ai et al. (2024)
conclude that there is a “limitation in LLMs’ ability to authentically replicate
human personality dynamics, often reflecting a bias towards socially desirable
responses” and \textcite{huQuantifyingPersonaEffect2024} conclude that
silicon-style “persona prompting cannot reliably simulate different
perspectives”. Likewise, we conclude that LLMs are not yet ready to fully
simulate individual-level human behaviours.

\printbibliography

\appendix
\section{Variables and transformations used to create silicon persona descriptions from the variables in the BBC dataset}
\label{app:variable_transformations_bbc}

Table \ref{tab:bbc_variables} lists the variables used from the BBC personality dataset,
descriptions, possible values and how these were used in the silicon persona
descriptions as part of the prompts.

\begin{landscape}
    \begin{table}[ht]
        \caption{Detailed description of variables used in the prompt}
        \label{tab:bbc_variables}
        \scriptsize 
        \begin{tabularx}{\linewidth}{@{}>{\hsize=0.15\hsize}X >{\hsize=0.20\hsize}X >{\hsize=0.35\hsize}X >{\hsize=0.30\hsize}X@{}}
            \toprule
            \textbf{Variable Name} & \textbf{Variable Description} & \textbf{Possible Values} & \textbf{Example Sentence(s) Used in the Prompt} \\
            \midrule
            age & Age & Number between 18 and 99 & I am 26 years old. \\
            ethnic & Ethnic background & 'Asian/Asian British - Indian; Pakistani; Bangladeshi', 'Black/black British', 'East/south-east Asian', 'Middle Eastern', 'Mixed race - white and Asian/Asian British', 'Mixed race - white and black/black British', 'White' & My ethnic background is White. \\
            m\_schl & Mother’s highest level of education & 'Did not complete GCSE / CSE / O-Levels', 'Completed GCSE / CSE / O-Levels', 'Completed post-16 vocational course', 'A-Levels', 'Undergraduate degree', 'Postgraduate degree' & The highest level of formal schooling my \{mother/father\} completed was GCSE / CSE / O-levels. \\
            f\_schl & Father’s highest level of education & 'Did not complete GCSE / CSE / O-Levels', 'Completed GCSE / CSE / O-Levels', 'Completed post-16 vocational course', 'A-Levels', 'Undergraduate degree', 'Postgraduate degree' & My father completed A-Levels. \\
            n\_sib & Number of siblings & 0, 1, 2, 3, 4, 5, '6 or more' & I have more than 5 siblings. \\
            sex & Sex & 'male', 'female' & I am male. \\
            st\_pub & Whether the majority of education up until 18 years of age was in a state or a private school & 'state', 'private' & The majority of my education up to the age of 18 was in a state school. \\
            occ\_st & Occupational status & 'Still at school', 'At university', 'In full time employment', 'Part time employment', 'Self employed', 'Homemaker/full-time parent', 'Unemployed', 'Retired' & My occupational status can be defined as Unemployed. \\
            occ\_cat & Occupational category & 'Accounting/finance', 'Administration', 'Business development', 'Consultancy', 'Customer service', 'Education / training', '…' & I work in Administration. \\
            income & Income & 'Up to £9,999 per annum (£199 per week)', '£10,000 to £19,999 per annum (£200 to £389 per week)', '£20,000 to £29,999 per annum (£390 to £579 per week)', '£30,000 to £39,999 per annum (£580 to £769 per week)', '£40,000 to £49,999 per annum (£770 to £969 per week)', '£50,000 to £74,999 per annum (£970 to £1,449 per week)', '£75,000 or more per annum (£1,450 or more per week)' & I earn £30,000 to £39,999 per annum. \\
            rstat\_1 & Relationship status & 'yes', 'no' & I am currently not in an intimate relationship. \\
            chldrn & Number of children & 0, 1, 2, 3, 4, 5, '6 or more' & I have 3 children. \\
            \bottomrule
        \end{tabularx}
    \end{table}
\end{landscape}

\section{Psychological scale instruments used}
\label{app:psychological_scales}

\textit{BFI}. The Big-Five Inventory \autocite{johnBigFiveTraitTaxonomy1999} is
a 44-item instrument that measures personality across five dimensions:
Extraversion, Conscientiousness, Openness to Experience, Neuroticism, and
Agreebleness. Raters are asked to judge how much an item is characteristic of
them. The items are rated on a 5-point Likert scale, where 1 = “Disagree
strongly” and 5 = “Agree strongly”.

\textit{PANAS}. The Positive and Negative Affect Schedule (PANAS) is a 20-item
instrument that measures two affect dimensions: positive and negative
\autocite{watsonDevelopmentValidationBrief1988}. Raters are asked how much they
agree with statements such as “I generally feed excited” on a 5-point Likert
scale ranging from “very slightly or not at all” to “extremely”.

\textit{SSCS}. The Short Scale of Creative Self (SSCS) is a 11-item instrument
that measures two dimensions of creativity: creative self-efficacy and creative
personal identity \autocite{karwowskiItDoesnHurt2011}. Raters are asked to
evaluate how well statements such as “I think I am a creative person” describe
them on a 5-point Likert scale ranging from “definitely not” to “definitely
yes”.

\textit{BPAQ}. The Buss-Perry Aggression Questionnaire is a 29-item instrument
that measures four dimensions of aggression: physical, verbal, anger, hostility
\autocite{bussAggressionQuestionnaire1992}. Raters are asked to evaluate how
characteristic statement such as “Once in a while I can't control the urge to
strike another person” are of them on a 5-point Likert scale ranging from
“extremely uncharacteristic” to “extremely characteristic”.

\section{Confirmatory factor analyses on the BPAQ, SSCS and PANAS scales}
\label{app:cfa_bpaq_sscs_panas}

\begin{figure*}
    \centering
    \includegraphics
    [trim=0 0 0 0,clip, width=0.95\textwidth]
    {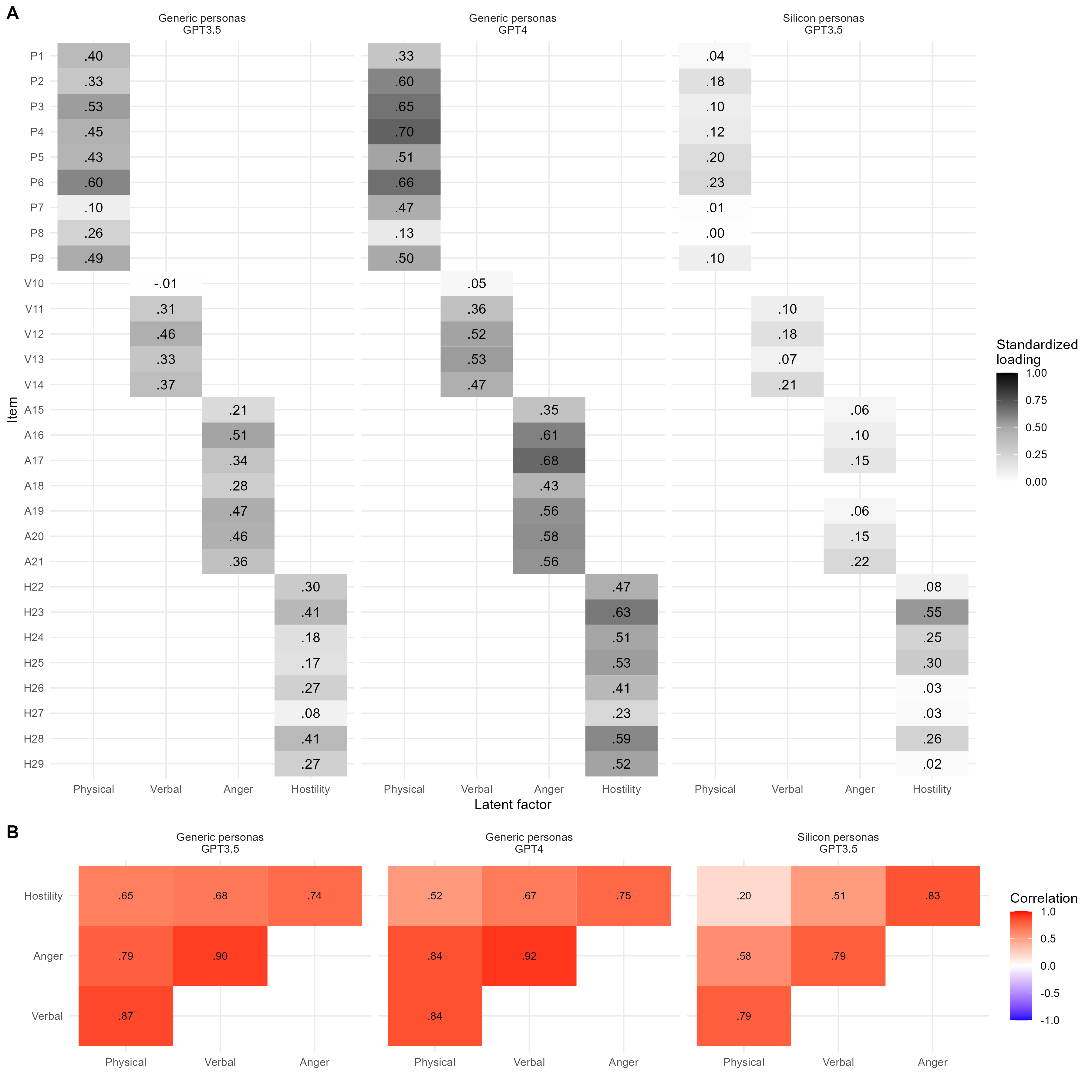}
    \caption{
        \small\textbf{Results from confirmatory factor analyses on the BPAQ scale across all models and prompting styles} Summary of the results from confirmatory factor analyses with the same model being fit on different data sources (columns). Panel A shows the path loadings from each observed variable (i.e. item, y-axis) to its latent factor (i.e. dimension, x-axis). Panel B shows the covariances between latent factors. NB: For the silicon personas, the responses of GPT-3.5 to items 10 (Verbal aggression) and 18 (Anger) were uniform and were thus excluded from model estimation.
    }
    \label{fig:cfa_results_bpaq}
\end{figure*}

\begin{figure*}
    \centering
    \includegraphics
    [trim=0 0 0 0,clip, width=0.95\textwidth]
    {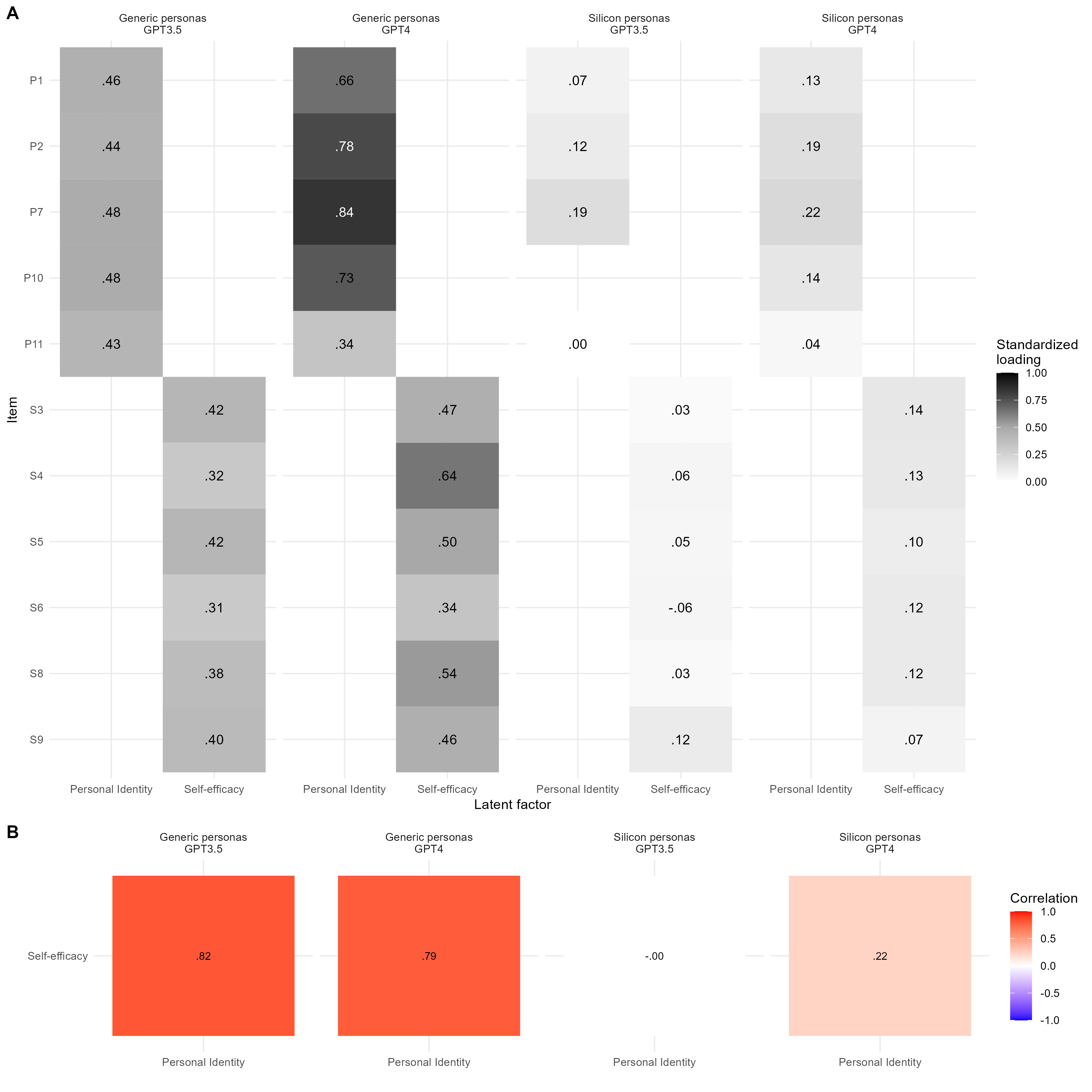}
    \caption{
        \small\textbf{Results from confirmatory factor analyses on the SSCS scale across all models and prompting styles} Summary of the results from confirmatory factor analyses with the same model being fit on different data sources (columns). Panel A shows the path loadings from each observed variable (i.e. item, y-axis) to its latent factor (i.e. dimension, x-axis). Panel B shows the covariances between latent factors. NB: For the silicon personas, the responses of GPT-3.5 to item 10 (Personal identity) was uniform and was thus excluded from the model estimation.
    }
    \label{fig:cfa_results_sscs}
\end{figure*}

\begin{figure*}
    \centering
    \includegraphics
    [trim=0 0 0 0,clip, width=0.95\textwidth]
    {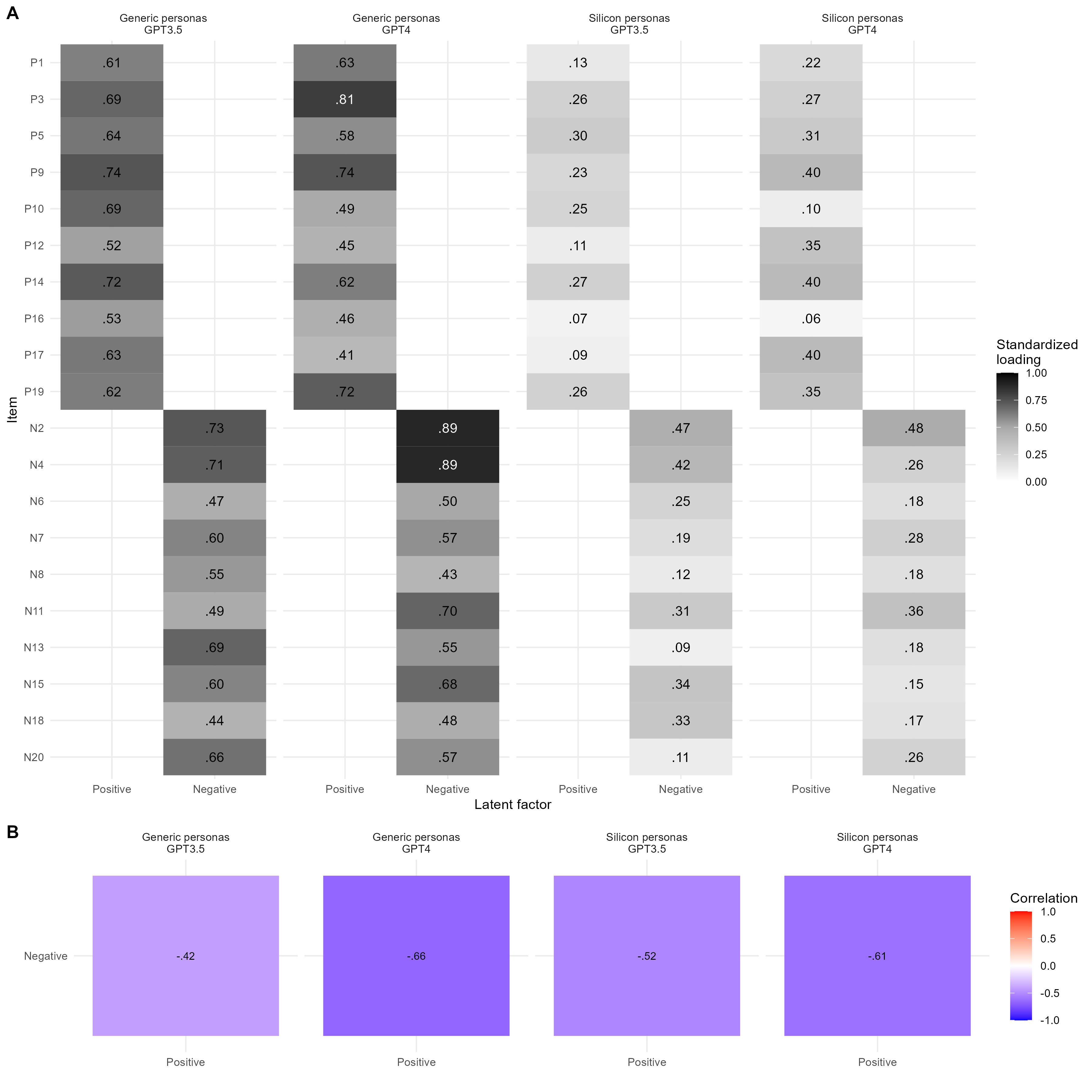}
    \caption{
        \small\textbf{Results from confirmatory factor analyses on the PANAS scale across all models and prompting styles} Summary of the results from confirmatory factor analyses with the same model being fit on different data sources (columns). Panel A shows the path loadings from each observed variable (i.e. item, y-axis) to its latent factor (i.e. dimension, x-axis). Panel B shows the covariances between latent factors.
    }
    \label{fig:cfa_results_panas}
\end{figure*}

We also ran confirmatory factor analyses on the BPAQ, SSCS and PANAS scales,
expecting to recover the factors for each scale (see Figure
\ref{fig:cfa_results_bpaq}, Figure \ref{fig:cfa_results_sscs}, Figure
\ref{fig:cfa_results_panas} and Table \ref{tab:bpaq_fit_indices}, Table
\ref{tab:sscs_fit_indices}, Table \ref{tab:panas_fit_indices}). As was the case with the
data from GPT-3.5 (both generic and silicon personas) on the BFI scale, here we
also witness a reliability paradox: the data from silicon personas (both GPT-3.5
and GPT-4) on all three scales, as well as data from GPT-3.5 using generic
personas for the BPAQ and SSCS scales show low path loadings ($\le$.40), with
fit indices indicating an otherwise good fit
\autocite{hancockReliabilityParadoxAssessing2011,mcneishThornyRelationMeasurement2018,milesTimePlaceIncremental2007}.
Thus, we conclude that the data from these LLMs and prompting variations do not
recover the expected factor structure.

As for the remaining variations, the data from GPT-4 using generic personas on
BPAQ and SSCS show decent, but ambiguous, loadings across the board, and their
fit indices (GFI=.72, IFI=.72, RMSEA=.09 and GFI=.82, IFI=.87, RMSEA=.14,
respectively) suggest a rather poor fit. The data from GPT-3.5 and GPT-4 on the
PANAS scale show high path loadings, however the model fit indices indicate a
tenuous fit (GFI=.82, IFI=.93, RMSEA=.08 and GFI=.69, IFI=.85, RMSEA=.11,
respectively). Thus, we conclude that data from LLMs do not have good structural
validity across the board, if at all.

\begin{table*}[ht]  
    \centering
    \caption{
        \small \textbf{Fit indices from the confirmatory factor analyses on the BPAQ scale across different data sources.}
        A confirmatory factor analysis using a maximum-likelihood estimation with robust standard errors was performed for each data source and 3 fit indices reported.
    }
    \label{tab:bpaq_fit_indices}
    \begin{tabularx}{\textwidth}{@{}Xccc@{}}  
        \toprule
        Data source               & GFI & IFI & RMSEA \\ 
        \midrule
        Generic sampling, GPT-3.5 & .78 & .85 & .06   \\
        Generic sampling, GPT-4   & .72 & .82 & .09   \\
        Silicon sampling, GPT-3.5 & .95 & .82 & .03   \\
        \bottomrule
    \end{tabularx}
\end{table*}

\begin{table*}[ht]  
    \centering
    \caption{
        \small \textbf{Fit indices from the confirmatory factor analyses on the SSCS scale across different data sources.}
        A confirmatory factor analysis using a maximum-likelihood estimation with robust standard errors was performed for each data source and 3 fit indices reported.
    }
    \label{tab:sscs_fit_indices}
    \begin{tabularx}{\textwidth}{@{}Xccc@{}}  
        \toprule
        Data source               & GFI & IFI & RMSEA \\ 
        \midrule
        Generic sampling, GPT-3.5 & .84 & .91 & .13   \\
        Generic sampling, GPT-4   & .82 & .87 & .14   \\
        Silicon sampling, GPT-3.5 & .95 & .81 & .08   \\
        Silicon sampling, GPT-4   & .84 & .58 & .15   \\
        \bottomrule
    \end{tabularx}
\end{table*}

\begin{table*}[ht]  
    \centering
    \caption{
        \small \textbf{Fit indices from the confirmatory factor analyses on the PANAS scale across different data sources.}
        A confirmatory factor analysis using a maximum-likelihood estimation with robust standard errors was performed for each data source and 3 fit indices reported.
    }
    \label{tab:panas_fit_indices}
    \begin{tabularx}{\textwidth}{@{}Xccc@{}}  
        \toprule
        Data source               & GFI & IFI & RMSEA \\ 
        \midrule
        Generic sampling, GPT-3.5 & .82 & .93 & .08   \\
        Generic sampling, GPT-4   & .69 & .85 & .11   \\
        Silicon sampling, GPT-3.5 & .88 & .83 & .08   \\
        Silicon sampling, GPT-4   & .91 & .85 & .05   \\
        \bottomrule
    \end{tabularx}
\end{table*}

\end{document}